\documentclass[conference]{IEEEtran}
\IEEEoverridecommandlockouts
\usepackage{cite}
\usepackage{amsmath,amssymb,amsfonts}
\usepackage{algorithmic}
\usepackage{graphicx}
\usepackage{textcomp}
\usepackage{xcolor}
\usepackage{tikz}

\usetikzlibrary{shapes.geometric,arrows,positioning}

\def\BibTeX{{\rm B\kern-.05em{\sc i\kern-.025em b}\kern-.08em
    T\kern-.1667em\lower.7ex\hbox{E}\kern-.125emX}}
\begin{document}

\title{Real-Time Imitation of Human Head Motions, Blinks and Emotions by
    Nao Robot: A Closed-Loop Approach}
\author{
    \IEEEauthorblockN{Keyhan Rayati}
    \IEEEauthorblockA{Human and Robot Interaction Lab,\\
        Electrical and Computer Engineering,\\
        University of Tehran,\\
        Tehran, Iran \\
        keyhan.rayati@ut.ac.ir}
    \and
    \IEEEauthorblockN{Amirhossein Feizi}
    \IEEEauthorblockA{Department of Electrical Engineering,\\
        Shahrood University of Technology,\\
        Shahrood, Iran \\
        amirhosseinfeizi68@gmail.com}
    \and
    \IEEEauthorblockN{Alireza Beigy}
    \IEEEauthorblockA{Human and Robot Interaction Lab,\\
        Electrical and Computer Engineering,\\
        University of Tehran,\\
        Tehran, Iran \\
        alireza.beigy@ut.ac.ir}
    \and
    \IEEEauthorblockN{Pourya Shahverdi}
    \IEEEauthorblockA{Intelligent Robotics Laboratory,\\
        Oakland University,\\
        Michigan, USA \\
        shahverdi@oakland.edu}
    \and
    \IEEEauthorblockN{Mehdi Tale Masouleh}
    \IEEEauthorblockA{Human and Robot Interaction Lab,\\
        Electrical and Computer Engineering,\\
        University of Tehran,\\
        Tehran, Iran \\
        m.t.masouleh@ut.ac.ir}
    \and
    \IEEEauthorblockN{Ahmad Kalhor}
    \IEEEauthorblockA{Human and Robot Interaction Lab,\\
        Electrical and Computer Engineering,\\
        University of Tehran,\\
        Tehran, Iran \\
        akalhor@ut.ac.ir}
    \and
    \IEEEauthorblockN{Wing-Yue Geoffrey Louie}
    \IEEEauthorblockA{Intelligent Robotics Laboratory,\\
        Oakland University,\\
        Michigan, USA \\
        louie@oakland.edu}
}

\maketitle

\begin{abstract}
    This paper introduces a novel approach for enabling real-time imitation of human head motion by a Nao robot, with a primary focus on elevating human-robot interactions.
    By using the robust capabilities of the MediaPipe as a computer vision library and the DeepFace as an emotion recognition library, this research endeavors to capture the subtleties of human head motion,
    including blink actions and emotional expressions, and seamlessly incorporate these indicators into the robot's responses.
    The result is a comprehensive framework which facilitates precise head imitation within human-robot interactions, utilizing a closed-loop approach that involves gathering real-time feedback from the robot's imitation performance.
    This feedback loop ensures a high degree of accuracy in modeling head motion, as evidenced by an impressive R2 score of 96.3 for pitch and 98.9 for yaw.
    Notably, the proposed approach holds promise in improving communication for children with autism, offering them a valuable tool for more effective interaction.
    In essence, proposed work explores the integration of real-time head imitation and real-time emotion recognition to enhance human-robot interactions, with potential benefits for individuals with unique communication needs.
\end{abstract}


\section{Introduction}
The field of robotics has come a long way in recent years, with significant advancements in the development of humanoid robots.
\begin{figure}[tp]
    \centering
    \includegraphics[width=\linewidth]{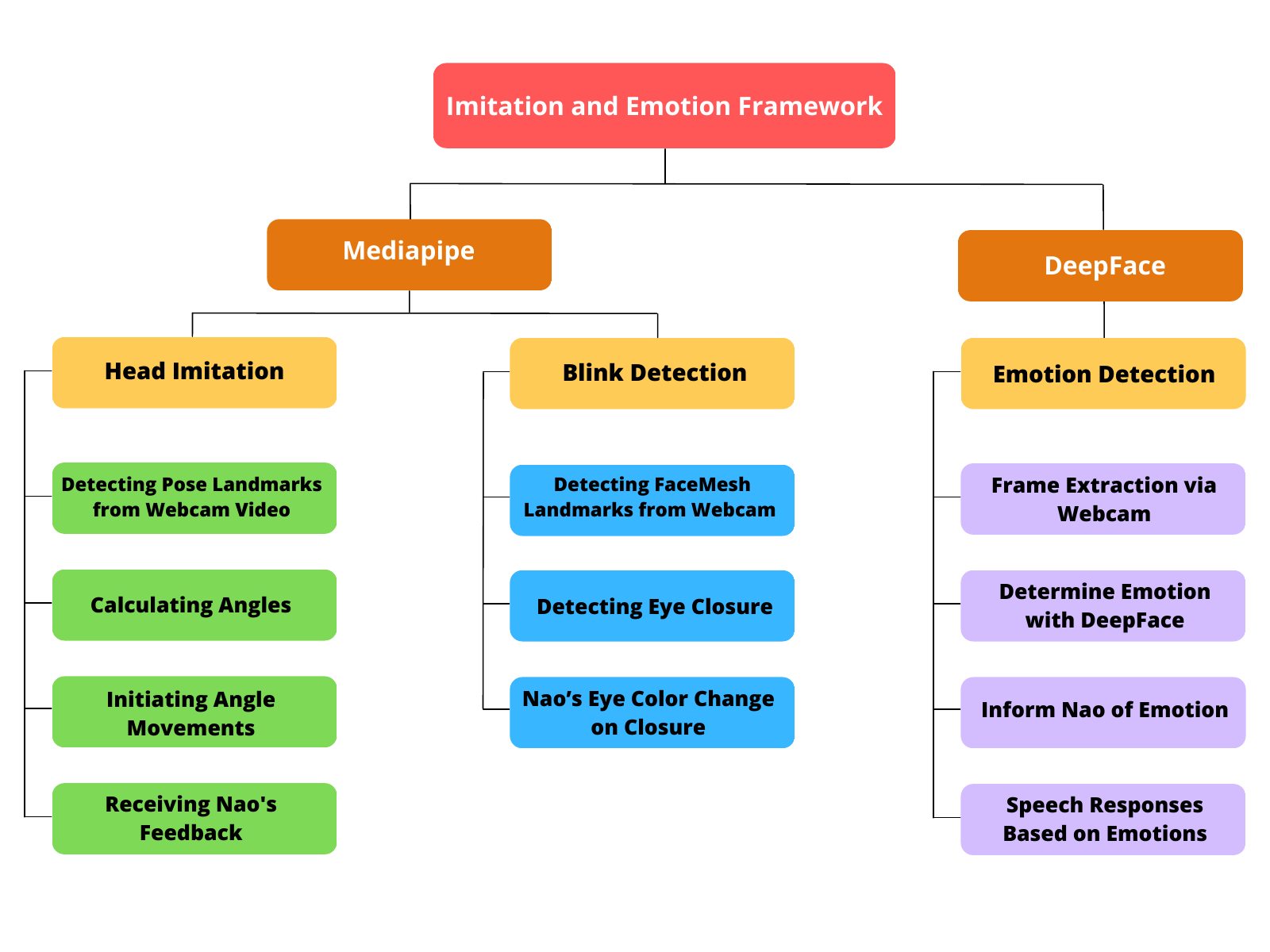}
    \caption{Framework overview: Real-time head motion imitation and emotion detection.}
    \label{fig:framework_structure}
\end{figure}
Humanoid robots are mostly designed to mimic human movement, behavior, and communication, making them ideal for a wide range of applications such as healthcare, education, entertainment, and more. In recent years, there has been a growing interest in imitating humanoid robots as researchers seek to improve the functionality and performance of these machines.
Robot imitation involves three primary stages, namely, observation, representation, and reproduction.
In the observation stage, Riley et al. \cite{b1} employed an advanced 3D vision system.
This system utilized both external cameras and head-mounted cameras to capture human movements meticulously.
The approach involved strategically placing colored markers on the human body, and subsequently, the precise positions of these markers were calculated and meticulously recorded.
Similar marker-based visual capture methods are documented in \cite{b2}.
Furthermore, wearable devices like "Xsens MVN" and "ShapeTape" have gained prominence as effective tools for capturing motion \cite{b3,b4,b5}.
Additionally, markerless visual capture systems, exemplified by the Microsoft Kinect sensor \cite{b6,b7,b8,b9,b10,b11}, offer a cost-effective and user-friendly alternative.
Such methods achieve robot imitation by harnessing the skeleton data provided by the Kinect.
Different techniques exist for translating human motions into movements for humanoid robots.
One strategy employs a geometrical analytical method grounded in link vectors within the robot's arms, hands, and head \cite{b12}.
Moreover, inverse kinematic motion models can also map human gestures onto robot actions.
These models incorporate the constraints and capabilities of the robot's joints and actuators, facilitating the generation of trajectories that are both precise and faithful to reality \cite{b1,b13,b16,b17,b18,b19}.

The approach proposed in this paper employs the robust capabilities of MediaPipe, a widely recognized and powerful computer vision framework.
For the purpose of this study, MediaPipe is used to capture and analyze the intricate details of human head motion, including pitch and yaw angles, providing the essential input for the robotic imitation process.
Leveraging the strengths of MediaPipe, this research extends its potential applications into the realm of human-robot interaction. Moreover, the proposed approach enhances the robot's ability to seamlessly mimic human head movements by incorporating blink detection using MediaPipe.
This additional feature enriches the capabilities of the proposed real-time imitation system and opens up new avenues for a more nuanced and interactive human-robot engagement \cite{b14}.
Alongside exploring MediaPipe's capabilities, the DeepFace\cite{b15} library is seamlessly integrated to make this research more comprehensive. DeepFace, a remarkable advancement in computer vision and facial recognition technology, represents a significant advancement in computer vision and facial recognition technology.
Developed by Facebook's AI Research (FAIR) team, it is a deep learning-based facial recognition system designed to excel in the challenging task of face verification.
DeepFace's core innovation lies in its ability to create a high-dimensional feature vector representation of facial images.
\begin{figure}[tp]
    \centering
    \begin{tikzpicture}[node distance=2cm]

        \tikzstyle{box} = [rectangle, rounded corners, minimum width=3cm, minimum height=1cm, text centered, draw=black, fill=blue!20]
        \tikzstyle{arrow} = [thick,->,>=stealth]

        \node (MediaPipe) [box] {MediaPipe};
        \node (framework) [box, below=of MediaPipe] {Proposed Framework};
        \node (naoqi) [box, right=of framework, xshift=1cm] {Naoqi API};

        \draw [arrow] (MediaPipe) -- node[anchor=west] {Data} (framework);
        \draw [arrow] (framework) to [out=0, in=180] node[above, midway] {HTTP Request} (naoqi);

    \end{tikzpicture}
    \caption{Interaction diagram of MediaPipe, Naoqi API, and framework.}
    \label{fig:interaction_diagram}
\end{figure}
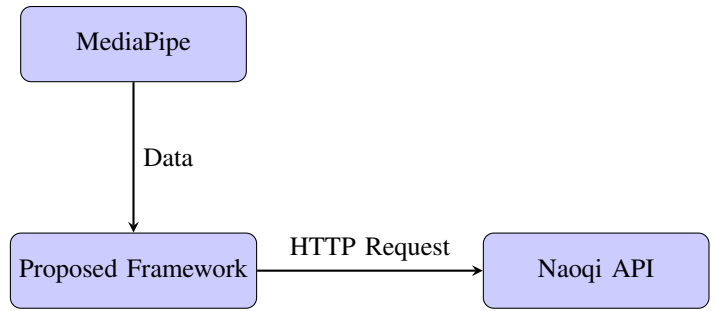
This feature vector enables it to distinguish between different faces with remarkable accuracy, rivaling human-level performance.
This breakthrough has paved the way for various applications in facial recognition, including the ability to detect and interpret human emotions, a capability which is integrated into the proposed approach in this paper for real-time imitation system of the Nao robot \cite{b15}.
In order to summarize, the contribution of this research revolves around achieving three key stages which are central to this work.
The primary goal consists of executing precise human head imitation through angle calculations.
The study focuses on pioneering real-time blink detection and imitation techniques.

Lastly, an emotion response system is introduced in order to enhance human-robot interactions.
As shown in Fig.~\ref{fig:framework_structure}, these three stages collectively form the essence of the comprehensive real-time imitation system, poised to revolutionize human-robot interaction.

The subsequent sections of this paper provide a detailed account of the three main phases of the research. In the following section, the implementation of the framework for head imitation,
blink detection, and emotion recognition is discussed. Afterward, the results of the comprehensive testing process for each component are presented.
Finally, the paper concludes by summarizing the findings and discussing their potential implications.

\section{Human Imitation and Interaction Framework}

In this section, a comprehensive framework for human imitation and interaction is presented, which leverages the capabilities of the MediaPipe for pose and face mesh detection, coupled with the Naoqi API.
The framework aims to analyze various human behaviors such as head movements, eye blinks, and emotions, and then imitate or respond to these behaviors using the Nao Robot.
The details of each component of this framework, challenges faced, and the strategies used to overcome them are discussed in the subsequent subsections.

\subsection{Environment Setup}

In this study, the MediaPipe version 0.10.2 is utilized.
The framework offers functionalities for both face mesh and pose detection.
The head rotation angles are determined based on eye and nose landmarks by a pose detection approach.
Also, the face-mesh capability is employed to discern the inner and outer eye landmarks to track blinking movements.

Fig.~\ref{fig:interaction_diagram} shows the data flow within the proposed framework.
This data comprises landmarks from pose and face mesh, sourced from MediaPipe, which are then processed within the proposed framework.
Then, the analyzed information including the yaw and pitch of the robot's head, eye status, and facial emotions is relayed to the Naoqi API. However, a challenge emerged during setup.
MediaPipe operates on Python 3.10, while the Naoqi API runs on Python 2.7.
In order to bridge this compatibility gap, some specific components are designed, which are integrated via HTTP requests. Additionally, it is noteworthy that the implementation leverages a simple camera or webcam, such as those found on a laptop, for data acquisition.
Importantly, this approach is extendable to any other standard camera, contributing to the accessibility and adaptability of the proposed framework.

\subsection{Head Imitation}
The head imitation component aims at computing the subject's head's yaw and pitch orientation.
By extracting face landmarks and plotting the landmarks, the eyes are observed to be aligned along the $y$-axis,
which means the eyes have the same $y$ and different $x$ and $z$.
\begin{figure}[tp]
    \centering
    \includegraphics[width=\linewidth]{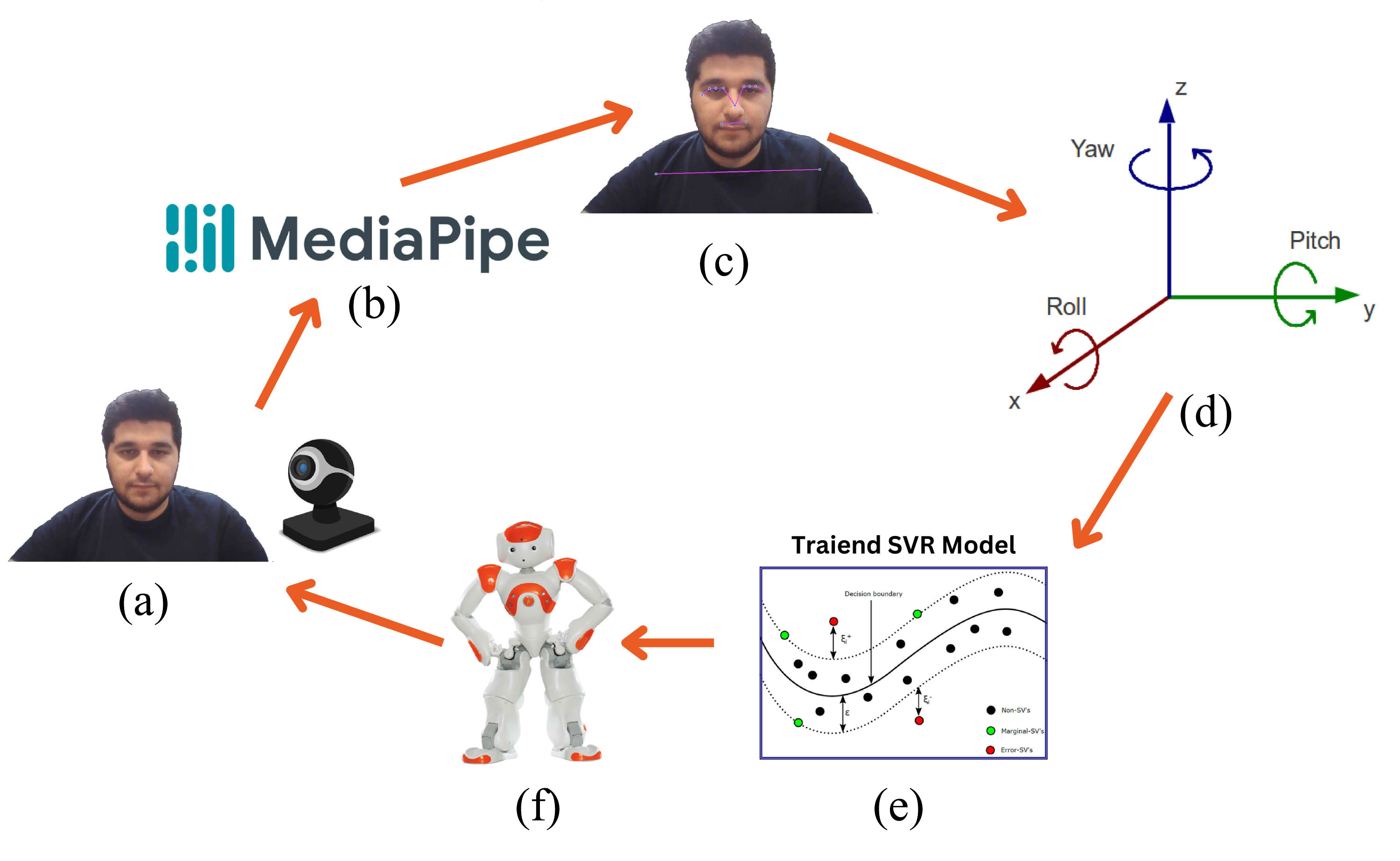}
    \caption{Real-Time Imitation System's Closed-Loop. (a) Capturing User's Face with Webcam, (b) Landmark Calculation Using MediaPipe, (c) Facial Landmarks Overlay, (d) Angle Calculations for Head Pitch and Yaw, (e) Predictive Angle Calculation for Unobserved Data, (f) Nao Robot Integration for Angle Execution, with Error Calculation Through Feedback.}
    \label{fig:diagram}
\end{figure}
\subsubsection{Head Yaw Estimation}

The estimation of head yaw is formulated as an angular measurement based on a predefined baseline vector and
the vector formed by the subject's landmarks corresponding to the left and right eyes.
Let \( \hat{\mathbf{V}}_{\text{baseline}} \) be the baseline vector corresponding to a neutral head pose.
Given the 3D coordinates of the subject's left and right eyes as \( \mathbf{P}_{\text{left}} \) and \( \mathbf{P}_{\text{right}} \), respectively, In the approach proposed by this paper, the eye-based vector \( \mathbf{V}_{\text{eye}} \) is calculated as follows:
\begin{align}
    \mathbf{V}_{\text{eye}} = \mathbf{P}_{\text{left}} - \mathbf{P}_{\text{right}}
\end{align}
The vector \( \mathbf{V}_{\text{eye}} \) is then normalized to a unit length:
\begin{align}
    \hat{\mathbf{V}}_{\text{eye}} = \frac{\mathbf{V}_{\text{eye}}}{\| \mathbf{V}_{\text{eye}} \|}
\end{align}
The angle \( \theta \) between \( \mathbf{V}_{\text{baseline}} \) and \( \hat{\mathbf{V}}_{\text{eye}} \) is obtained using the dot and cross products as:
\begin{align}
    \theta = \arccos\left( \hat{\mathbf{V}}_{\text{baseline}} \cdot \hat{\mathbf{V}}_{\text{eye}} \right)
\end{align}
The axis of rotation \( \mathbf{a} \) is calculated as follows:
\begin{align}
    \mathbf{a} = \frac{\hat{\mathbf{V}}_{\text{baseline}} \times \hat{\mathbf{V}}_{\text{eye}}}{\| \hat{\mathbf{V}}_{\text{baseline}} \times \hat{\mathbf{V}}_{\text{eye}} \| }
\end{align}
A rotation vector \( \mathbf{r} \) is then formed as:
\begin{align}
    \mathbf{r} = \theta\mathbf{a}
\end{align}
This rotation vector is converted into Euler angles to extract the yaw\( (\phi) \) component.
In order to ensure robustness in real-world scenarios, \( \phi \) is capped within the range \([-119.5^\circ, 119.5^\circ]\).

\subsubsection{Head Pitch Estimation}

The head's pitch is determined based on the vector's orientation formed by the eyes' mid-point and the nose relative to a baseline vector corresponding to a neutral head pose.
Let \( \hat{\mathbf{V}}_{\text{baseline}} \) be the baseline vector, normalized to unit length. The 3D coordinates for the mid-point between the subject's left and right eyes \( \mathbf{M}_{\text{eye}} \) and the subject's nose \( \mathbf{P}_{\text{nose}} \) are given.
Firstly, \( \mathbf{M}_{\text{eye}} \) is calculated as the mean of the left and right eye coordinates:
\begin{align}
    \mathbf{M}_{\text{eye}} = \frac{\mathbf{P}_{\text{left}} + \mathbf{P}_{\text{right}}}{2}
\end{align}
The vector \( \hat{\mathbf{V}}_{\text{new}} \) from the mid-point of the eyes to the nose is then formulated as:
\begin{align}
    \hat{\mathbf{V}}_{\text{new}} = \mathbf{P}_{\text{nose}} - \mathbf{M}_{\text{eye}}
\end{align}
This vector is normalized to a unit length:
\begin{align}
    \hat{\mathbf{V}}_{\text{new}} = \frac{\mathbf{V}_{\text{new}}}{\| \mathbf{V}_{\text{new}} \|}
\end{align}
The angle \( \alpha \) between \( \hat{\mathbf{V}}_{\text{baseline}} \) and \( \hat{\mathbf{V}}_{\text{new}} \) is calculated using the dot product as:
\begin{align}
    \alpha = \arccos\left( \hat{\mathbf{V}}_{\text{baseline}} \cdot \hat{\mathbf{V}}_{\text{new}} \right)
\end{align}
The axis of rotation \( \mathbf{a} \) is calculated using the cross product:
\begin{align}
    \mathbf{a} = \frac{\hat{\mathbf{V}}_{\text{baseline}} \times \hat{\mathbf{V}}_{\text{new}}}{\| \hat{\mathbf{V}}_{\text{baseline}} \times \hat{\mathbf{V}}_{\text{new}} \| }
\end{align}
A rotation vector \( \mathbf{r} \) is then formed as:
\begin{align}
    \mathbf{r}=\alpha\mathbf{a}
\end{align}
This rotation vector is converted into Euler angles, to extract the pitch\( (\psi) \) component.
The allowable pitch angles are dependent on the current yaw angle value.
To predict rotation angles within specified collision limits, a Support Vector Regression (SVR) model is trained using known values provided by the Nao robot manufacturer. This training enables the model to learn a mapping function, which predicts rotation angles for scenarios where exact values are unknown or fall between the provided training data, ensuring stability and preventing movements beyond the robot's defined collision thresholds.
For the sake of enhancing the robustness of the estimate against outliers, \( \psi \) is confined within a range \( [\text{min}, \text{max}] \). The boundary conditions for \( \psi \) are defined as:
\begin{align}
    \text{min} = SVR_{\text{min\_pitch}}(\phi) + 0.05 \times SVR_{\text{min\_pitch}}(\phi)
\end{align}
\begin{align}
    \text{max} = SVR_{\text{max\_pitch}}(\phi) - 0.05 \times SVR_{\text{max\_pitch}}(\phi)
\end{align}
The final pitch value is defined as:
\begin{align}
    \psi = \begin{cases}
               \text{max} & \text{if } \psi > \text{max} \\
               \text{min} & \text{if } \psi < \text{min} \\
               \psi       & \text{otherwise}
           \end{cases}
\end{align}
\begin{figure}[tp]
    \centering
    \includegraphics[width=\linewidth]{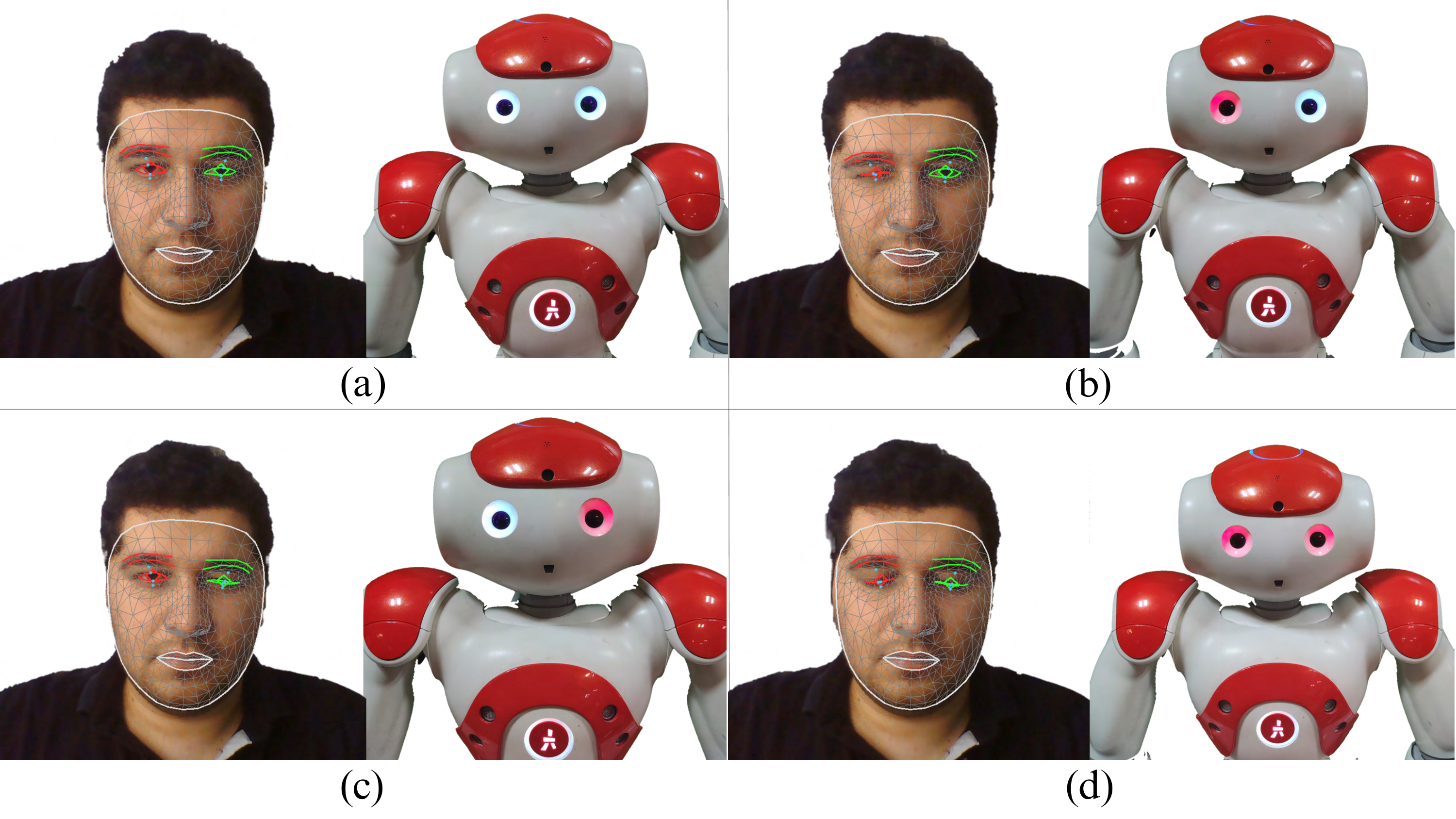}
    \caption{Eye closure interactions, (a) Both eyes open, (b) Right eye closed, (c) Left eye closed and (d) Both eyes closed.}
    \label{fig:blink_interactions}
\end{figure}
After this initial calculation of angles, the proposed framework enters a closed-loop system, as depicted in Fig.~\ref{fig:diagram}.
In this closed-loop configuration, the observed head motions are not only commanded to be replicated by the Nao robot, but feedback is also continuously received from the robot's sensors, which monitor the accuracy and synchronization of the imitation.
This feedback mechanism plays a crucial role in refining the precision of the proposed real-time imitation, ensuring that the robot's head movements closely mirror those of the human operator.

\subsection{Blink Detection}
In order to improve the human imitation system, a blink detection framework is introduced, which determines the open or closed state of a subject's eyes.
Using the face mesh model provided by MediaPipe, a geometry-based approach is employed to estimate the eye status.
At least, blink estimation is applied to the Nao Robot, as shown in Fig.~\ref{fig:blink_interactions}

\subsubsection{Defining Eye Regions}
Analyzing the face mesh, the distances between the top and bottom points of the eye's inner and outer sections are computed.
these distances are denoted as \( d_{\text{inner}} \) and \( d_{\text{outer}} \), respectively:
\begin{align}
    d_{\text{inner}} = || it - ib ||
\end{align}
\begin{align}
    d_{\text{outer}} = || ot - ob ||
\end{align}
where the \( it \), \( ib \), \( ot \), and \( ob \) stand for the inner top, inner bottom, outer top, and outer bottom eye points, respectively.
Finally, the eye ratio \( R \), which will be used for defining the blinking threshold, can calculated as follows:

\begin{align}
    R = \frac{|d_{\text{outer}}|}{|d_{\text{inner}}|}
\end{align}

\subsubsection{Blink Thresholding}
A blink is detected when the computed eye ratio \( R \) is higher than a predefined threshold \( T \).
This threshold is empirically determined because of differences in eye shape:
\begin{align}
    \text{Eye status} =
    \begin{cases}
        \text{Closed} & \text{if } R > T \\
        \text{Open}   & \text{otherwise}
    \end{cases}
\end{align}
This approach is applied independently to both eyes, allowing for the identification of blinks for each eye.

\begin{figure}[tp]
    \centering
    \includegraphics[width=\linewidth]{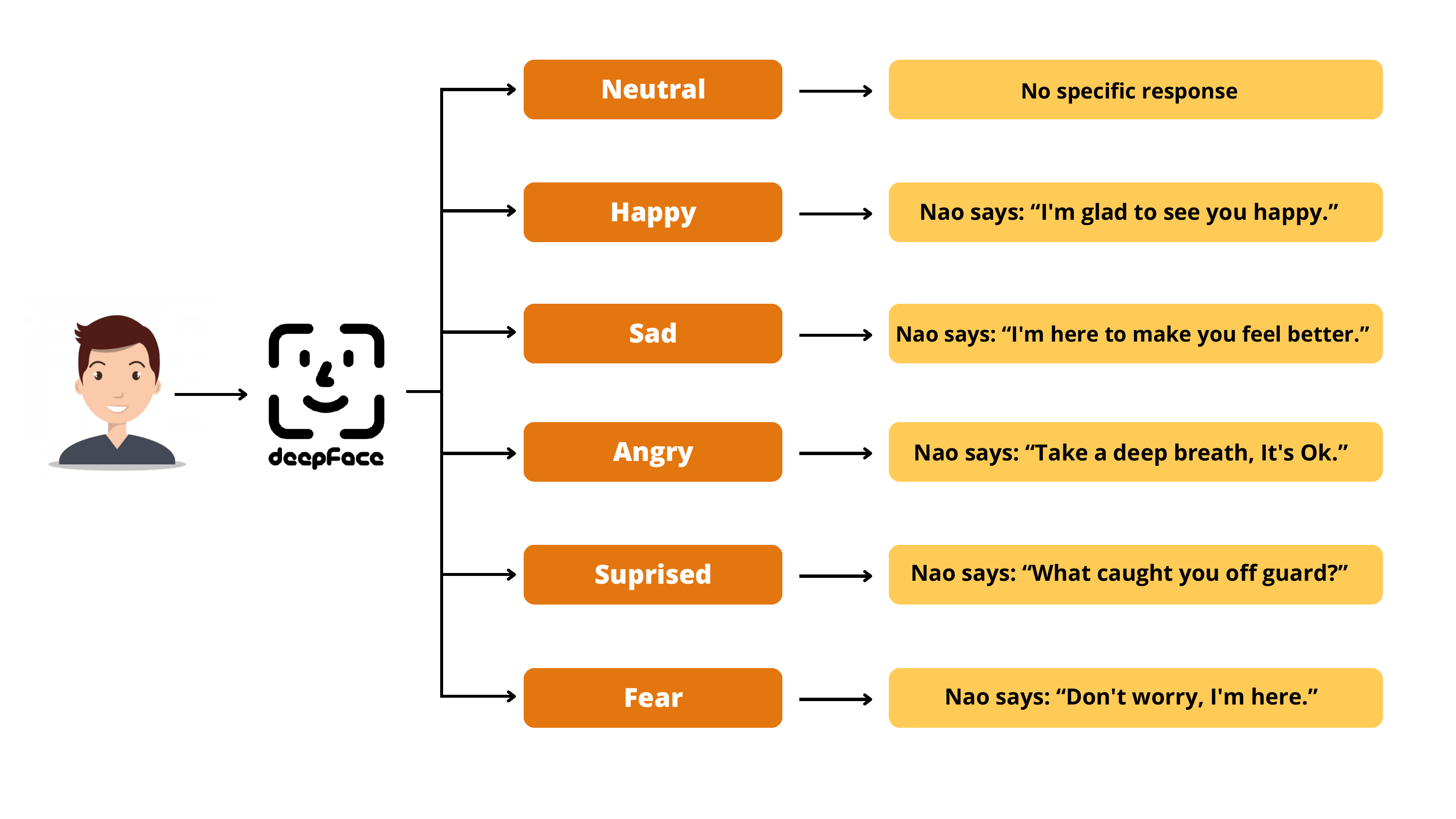}
    \caption{Emotion-to-Response Mapping for the Nao robot using DeepFace analysis.}
    \label{fig:emotion_mapping}
\end{figure}
\subsection{Emotion Detection}
The DeepFace library is seamlessly integrated into the proposed framework, employing it for real-time facial emotion detection.
The webcam images are analyzed frame by frame, ensuring precision in the observations.
In order to minimize noise and errors in the detection process, emotions over a sequence of 10 frames were collected, and then a majority feelings approach was adopted.
As depicted in Figure \ref{fig:emotion_mapping}, the method allowed to effectively identify the prevailing emotions, encompassing a wide range of human expressions, including anger, fear, neutrality, sadness, disgust, happiness, and surprise.
These detected emotions are subsequently associated with prepared textual responses for the Nao robot.
Leveraging the robot's text-to-speech capabilities, the robot dynamically communicated these responses, enhancing its capacity to engage with users in real-time, tailored to their evolving emotional states.
By dynamically responding to emotional signals, this innovative approach could potentially assist children with autism in enhancing their interactive experiences.
Tailoring the robot's responses to their evolving emotional states, opens up the possibility of more effective and engaging interactions, providing valuable support and opportunities for these children.

\section{Results and Experiments}
This section presents the results from the performed experiment, which was concerned with a Nao robot's ability to mimic human head motion in real-time accurately.
Several experiments were run to assess the precision and efficacy of proposed strategy.

\subsection*{Test 1: Head Motion Imitation}
\begin{figure}[tp]
    \centering
    \includegraphics[width=\linewidth]{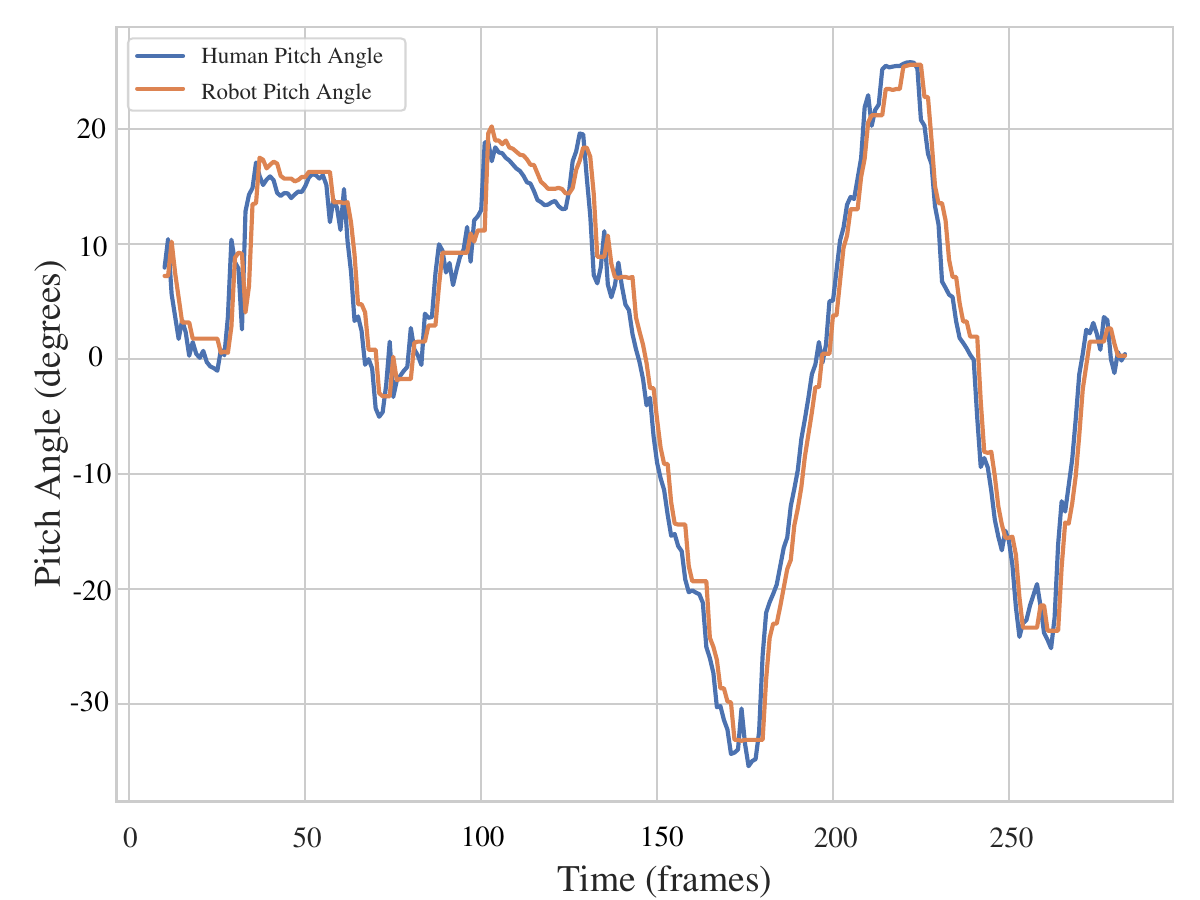}
    \caption{Comparison of human and robot pitch angles.}
    \label{fig:pitch_angles_comparison}
\end{figure}

In initial experiment, how well the robot's head movements replicated human subject motions was evaluated. The procedure involved the following steps:
\begin{enumerate}
    \item{Data Collection:} Head motion data is captured by manually moving the human head and recording the robot's head angles simultaneously.
    \item{Angle Calculation:} Using proposed method, the yaw and pitch angles for the human subject are computed.
    \item{Data Comparison:} The calculated angles to assess the degree of imitation by the robot are compared.
\end{enumerate}
\begin{figure}[tp]
    \centering
    \includegraphics[width=\linewidth]{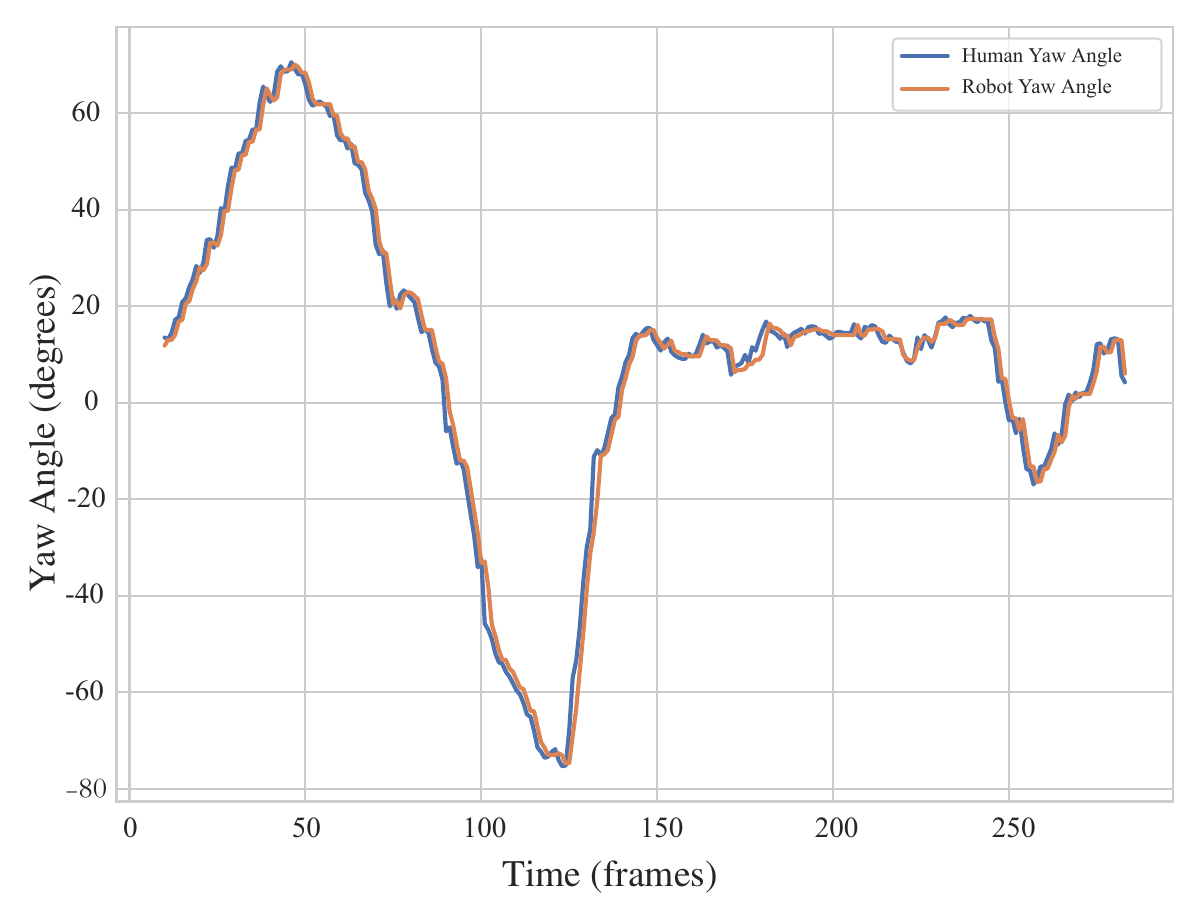}
    \caption{Comparison of human and robot yaw angles.}
    \label{fig:yaw_angles_comparison}
\end{figure}
It can be observed from Fig.~\ref{fig:yaw_angles_comparison} that the plot illustrates the yaw angles of the human head and the robot's head over time, enabling a visual comparison of their movements. In order to assess the accuracy of imitation, the R-squared (R²) value was measured, which yielded an impressive accuracy score of 98.9 for yaw angles.
\begin{figure}[tp]
    \centering
    \includegraphics[width=\linewidth]{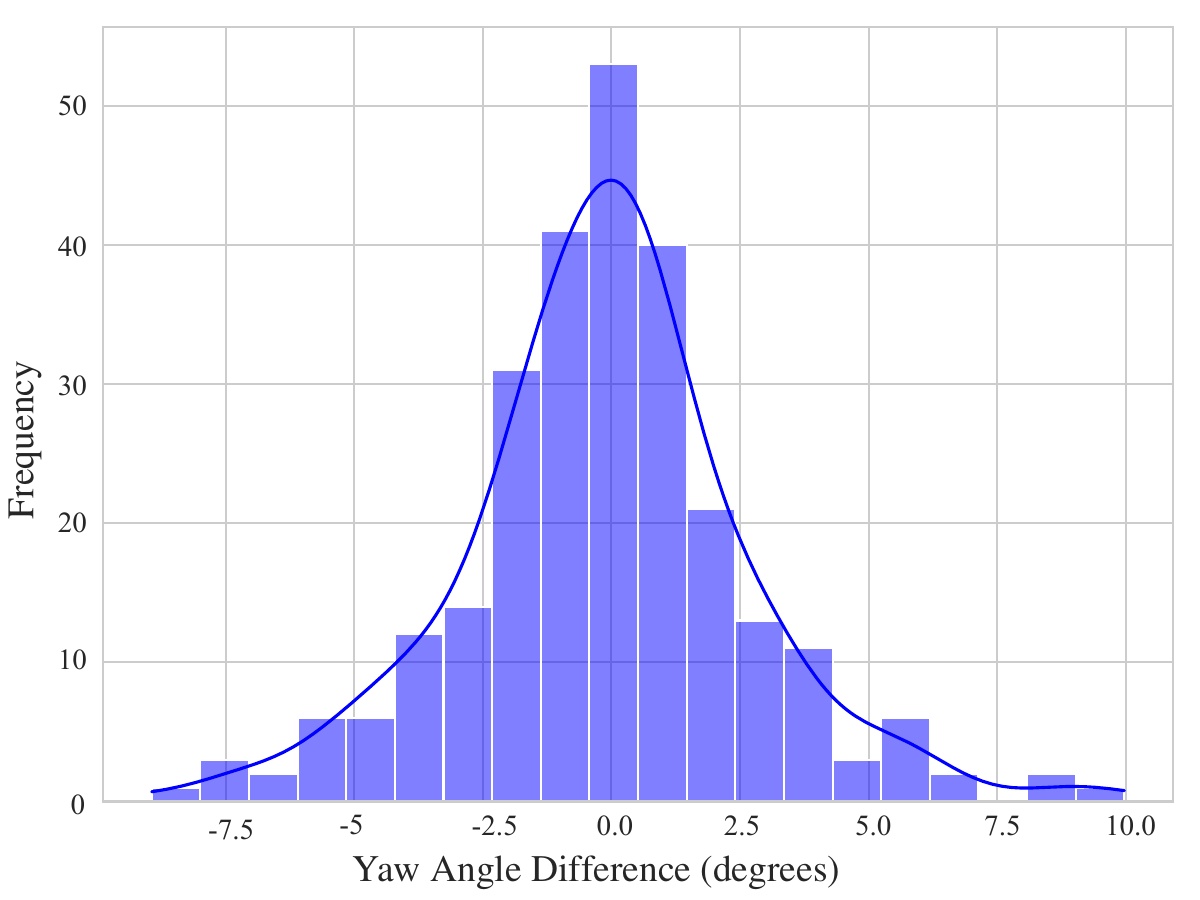}
    \caption{Distribution of yaw angle differences.}
    \label{fig:distribution_of_yaw_angle_differences}
\end{figure}
Fig.~\ref{fig:distribution_of_yaw_angle_differences} depicts the distribution of differences in yaw angles between the human and robot heads over time, serving as a measure of imitation accuracy.
In Fig.~\ref{fig:pitch_angles_comparison}, the pitch angles of the human head and the robot's head are illustrated, facilitating a visual assessment of their alignment. The distribution of both errors is calculated via ~250 frames of the experiment we had.

\begin{figure}[tp]
    \centering
    \includegraphics[width=\linewidth]{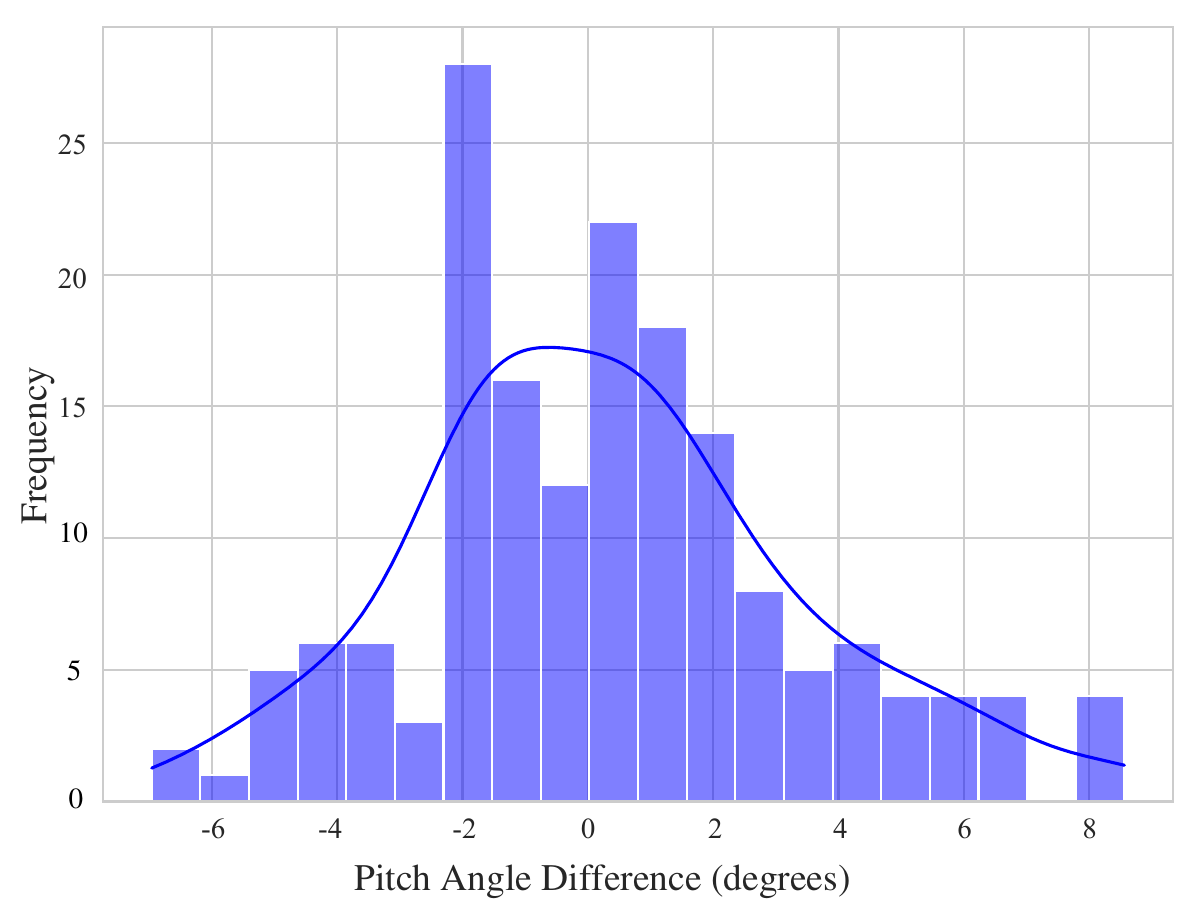}
    \caption{Distribution of pitch angle differences.}
    \label{fig:distribution_of_pitch_angle_differences}
\end{figure}

The distribution of differences in pitch angles between the human and robot heads over time reveals compelling insights into the accuracy of the proposed imitation method, which can be observed from Fig.~\ref{fig:distribution_of_pitch_angle_differences}.
In order to assess the accuracy of imitation, we measured the R-squared (R²) value, which yielded an impressive accuracy score of 98.9 for pitch angles.
Performed analysis reveals that the majority of pitch angle differences fall within a range of -6 to +8 degrees.
This range signifies that the robot's imitation closely mirrors the pitch movements of the human head with an acceptable level of variance.
Furthermore, the distribution exhibits a near-normal distribution, with a noteworthy observation.
The average pitch angle difference between the human and robot heads is approximately -2 degrees, indicating a slight offset of about 2 degrees from perfect alignment.
This offset, while small, suggests a consistent trend in the robot's pitch angle behavior that can be further explored and potentially calibrated for even greater accuracy.

\subsection*{Test 2: Blink Imitation}
A test focused on blink actions was conducted further to evaluate the imitation capabilities of the Nao robot.
The experiment proceeded as follows:

\begin{enumerate}
    \item{Data Collection:} Utilizing a webcam, a series of blinks for a total of 50 occurrences was initiated.
    Throughout these blink actions, the robot's response was closely monitored to ascertain if it imitated them.
    \item{Results:} In 48 out of the 50 blink occurrences, the robot exhibited a corresponding blink action, which exhibits its capability to imitate effectively human blinks in real-time.
    \item{Noise Handling:} The 48 successful blink imitations were attributed to the precision of the proposed method in detecting and replicating blink actions initiated by the human subject.
    However, it is noteworthy that the remaining 2 blink attempts failed due to their extremely brief duration, lasting less than 2 frames.
    In these instances, the rapid eye movement was considered noise by the Imitation approach of the Nao Robot, leading to a non-triggering of the robot's blink response.
\end{enumerate}
Test 2 demonstrated the Nao robot's ability to replicate blink actions initiated by a human subject accurately.
In order to further evaluate the imitation capabilities of the Nao robot, a test focused on blink actions was conducted, and the test with an additional four participants was replicated to ensure the consistency of the findings. The experiment proceeded as follows.
In addition, the same experiment with four more participants was conducted, and these subsequent tests yielded similar results.
This consistency across multiple trials further shows the robustness and reliability of the framework.
The high success rate of the proposed blink imitations algorithm, achieved through proposed method's precision, underscores the reliability and real-time capabilities of the framework.
This experiment confirms the effectiveness of the proposed approach in imitating blink actions, a crucial aspect of human head motion imitation.
The results from Test 2 further support the accuracy and real-time performance of the Imitation approach of the Nao Robot.

\subsection*{Test 3: Emotion Detection and Response}
In this experiment, the Nao robot's ability to detect and respond to human emotions in real-time. The methodology involved the following steps:
\begin{enumerate}
    \item{Emotion Expression:} A human subject was positioned in front of the webcam and prompted to express a range of emotions, including happiness, sadness, anger, surprise, and fear  through natural facial expressions.
    \item{Emotion Detection:} Real-time emotion detection was facilitated by the DeepFace library, enabling precise discernment and quantification of the human subject's emotions at any given moment.
    \item{Robot Response:} The Nao robot, equipped with text-to-speech capabilities, continuously analyzed the human subject's emotional state.
    The robot formulated and delivered contextually tailored verbal responses based on the detected emotions.
\end{enumerate}
Test 3 yielded highly promising outcomes, affirming the Nao robot's remarkable real-time emotion detection and response proficiency.
Notably, the robot's recognition of expressions of happiness was a standout feature.
Beyond happiness, the Nao robot effectively identified and responded to a spectrum of other emotions, encompassing sadness, anger, surprise, and fear.
Moreover, the robot's responses were contextually relevant and attuned to the expressed emotions, enhancing the overall quality of human-robot interaction.

In summary, from the performed experimentation, it can be inferred that the Nao robot, equipped with the proposed real-time imitation approach, excels in multiple facets of human-robot interaction.
Across a series of tests encompassing head motion imitation, blink actions, and emotion detection and response, the Nao robot consistently showcased remarkable accuracy and adaptability.
Notably, it accurately emulated human head motions, aligning closely with yaw and pitch angles.
Additionally, the robot's capability to imitate blink actions, particularly in real-world scenarios, was highly reliable.
Furthermore, the robot exhibited a deep understanding of human emotions, with the ability to detect and respond to a diverse range of emotional expressions.
These findings collectively affirm the effectiveness and versatility of the framework, paving the way for enhanced applications in the realm of human-robot interaction.

\section{Conclusions}
This research has embarked on a journey to enhance human-robot interaction through a comprehensive framework designed for real-time imitation of human head motion by a Nao robot.
This journey unfolded across three pivotal phases, each contributing to realizing a versatile and adaptable system.
In the initial phase, a novel method was proposed utilizing captured webcam video frames and the robust capabilities of the MediaPipe library.
This approach allowed to precisely calculate and replicate head motion angles in real-time, effectively closing the loop between the human operator and the Nao robot.
The resulting system exhibited exceptional accuracy, providing valuable insights into the robot's imitation capabilities.
The second phase focused on the intricacies of human blink actions.
Leveraging Mediapipe's landmark tracking, a technique to detect and replicate blink movements initiated by the human subject was developed.
The outcomes were resoundingly successful, highlighting the reliability of the proposed method in replicating real-world scenarios and underscoring the potential applications in human-robot interaction.
In the third and final phase, the realm of human emotion tracking was delved into. Utilizing the capabilities of Deepface, the Nao robot was able to interpret human emotions and respond accordingly.
This transformative addition to the proposed framework opened up new horizons for interactive experiences, with the robot adeptly responding to a spectrum of emotions, including happiness, sadness, anger, surprise, and fear.
These three pivotal phases collectively culminated in creating a complete head imitation framework.
This framework offers researchers and robot enthusiasts a powerful tool that seamlessly integrates head motion imitation, blink replication, and emotion recognition, all in real-time. While a broad range of potential applications is envisioned, one particularly notable avenue is the potential benefit to children with autism.
The framework has the capacity to enhance their interactive experiences, fostering more effective communication.

\bibliographystyle{IEEEtran}
\bibliography{references}
\vspace{12pt}

\end{document}